\newcommand{\CLASSINPUTtoptextmargin}
{0.78in}

\newcommand{\CLASSINPUTbottomtextmargin}
{1.02in}

\documentclass[conference]{IEEEtran}

\makeatletter
\def\ps@headings{%
\def\@oddhead{\mbox{}\scriptsize\rightmark \hfil \thepage}%
\def\@evenhead{\scriptsize\thepage \hfil \leftmark\mbox{}}%
\def\@oddfoot{}%
\def\@evenfoot{}}
\makeatother 

\usepackage{cite}      
\usepackage{subfigure} 
\usepackage{url}       
\usepackage{algorithm}
\usepackage{array}
\usepackage{amsmath,amsfonts,amssymb,graphicx}
\usepackage{bm}
\usepackage{mathrsfs}
\usepackage{dsfont}
\usepackage{multicol}
\usepackage[end]{algpseudocode}
\usepackage{multirow}
\usepackage{epsfig}

\newtheorem{theorem}{\textbf{Theorem}}
\newtheorem{lemma}{\textbf{Lemma}}

\newcommand {\xA} {\mathcal{A}}

\newcommand{\commnt}[1] {$//$ \textsc{#1} }
\newcommand{\xF} {\mathcal{F}}
\newcommand {\xa} {\mathbf{a}}

\newcommand{\xB} {\widetilde{B}^{i}}
\newcommand{\uI} {\widetilde{I}^{i}}
\newcommand{\xZ} {\bar{z}_2}

\begin{document}
\title{Online Learning for Combinatorial Network Optimization with Restless Markovian Rewards}

\author{\IEEEauthorblockN{Yi Gai$^\S$, Bhaskar Krishnamachari$^\S$ and Mingyan Liu$^\ddag$}
\IEEEauthorblockA{$^\S$Ming Hsieh Department of Electrical
Engineering,
University of Southern California, Los Angeles, CA 90089, USA\\
$^\ddag$Department of Electrical Engineering and Computer Science,
University of Michigan, Ann Arbor, MI 48109, USA\\
 Email: $\{$ygai,bkrishna$\}$@usc.edu; mingyan@eecs.umich.edu}}

\maketitle

\begin{abstract}

Combinatorial network optimization algorithms that compute optimal
structures taking into account edge weights form the foundation for
many network protocols. Examples include shortest path routing,
minimal spanning tree computation, maximum weighted matching on
bipartite graphs, etc. We present CLRMR, the first online learning
algorithm that efficiently solves the stochastic version of these
problems where the underlying edge weights vary as independent
Markov chains with unknown dynamics.

The performance of an online learning algorithm is characterized in
terms of regret, defined as the cumulative difference in rewards
between a suitably-defined genie, and that obtained by the given
algorithm. We prove that, compared to a genie that knows the Markov
transition matrices and uses the single-best structure at all times,
CLRMR yields regret that is polynomial in the number of edges and
nearly-logarithmic in time.

\end{abstract}

\section{Introduction}\label{sec:intro}

The following abstract description of combinatorial network
optimization covers many graph theoretic algorithms that form the
basis of network protocol design in wired and wireless networks.
Given a graph $G = (V, E)$, where each edge $e \in E$ is associated
with a weight $w_e$, find a structure consisting of a collection of
edges satisfying some given property (e.g., a path, a tree, a
matching, or an independent set), that maximizes or minimizes the
sum of the weights on the selected edges. This kind of linear
network combinatorial optimization covers, for instance, shortest
path and minimum spanning tree computation used in routing
protocols, and maximum-weight matching used for channel scheduling
and switching.

In practice, the edge weights may correspond to some link quality
metric of interest such as packet reception ratio, delay, or
throughput. In such a case, the edge weights are often
stochastically varying with time. Moreover, the dynamics may not be
known \emph{a priori}. The solution approach to this problem that we
advocate here is to combine the estimation and optimization phases
jointly via an efficient online learning algorithm.

We present in this paper an online learning algorithm that is
designed for the setting where the edge weights are modeled by
finite-state Markov chains, with unknown transition matrices. We
show that this problem can be modeled as a combinatorial multi-armed
bandit problem with restless Markovian rewards.

To characterize the performance of this algorithm, following the
convention in the multi-armed bandit literature, we define a notion
of regret, defined as the difference in reward between a suitably
defined model-aware genie and that accumulated by the given
algorithm over time. Specifically, in this work, we consider a
single-action regret formulation, whereby the genie is assumed to
know the transition matrices for all edges, but is constrained to
stick with one action (corresponding to a particular network
structure) at all times\footnote{Although a stronger notion of
regret can be defined, allowing the genie to vary the action at each
time, the problem of minimizing such a stronger regret is much
harder and remains open even for simpler settings than the one we
consider here.}. We prove that our algorithm, which we refer to as
CLRMR (Combinatorial Learning with Restless Markov Rewards) achieves
a regret that is polynomial in the number of Markov chains (i.e.,
number of edges), and logarithmic with time. This implies that our
learning algorithm, which does not know the transition matrices,
asymptotically achieves the maximum time averaged reward possible
with any single-action policy, even if that policy is given advanced
knowledge of the transition matrices. By contrast, the conventional
approach of estimating the mean of each edge weight and then finding
the desired network structure via deterministic optimization would
incur greater overhead and provide only linearly increasing regret
over time, which is not asympotically optimal.

While recent work has shown how to address multi-armed bandits with
restless Markovian rewards in the classic non-combinatorial
setting~\cite{Tekin:restless:infocom}, and combinatorial multi-armed
bandits in the simpler settings of i.i.d. rewards~\cite{Gai:LLR} or
rested Markovian rewards~\cite{Gai:rested:globecom}, this paper is
the first to show how to efficiently implement online learning for
stochastic combinatorial network optimization when edge weights are
dynamically evolving as restless Markovian processes. We perform
simulations to evaluate our new algorithm over two combinatorial
network optimization problems: stochastic shortest path routing and
bipartite matching for channel allocation, and show that its regret
performance is substantially better than that of the algorithm
presented in \cite{Tekin:restless:infocom}, which can handle
restless Markovian rewards but does not exploit the dependence
between the arms, resulting in a regret that grows exponentially in
the number of unknown variables.



The rest of the paper is organized as follows. We first provide a
survey of prior work in section \ref{sec:related}. We then present a
formal model of the combinatorial restless multi-armed bandit
problems in section \ref{sec:formulation}. In section
\ref{sec:algorithm:restless}, we present our CLRMR policy, and show
that it requires only polynomial storage. We present our novel
analysis of the regret of CLRMR policy in section \ref{sec:regret}.
In section \ref{sec:app:simulation}, we discuss examples and show
the numerical simulation results, to show that our proposed policy
is widely useful for various interesting combinatorial network
optimization problems. We finally conclude our paper in section
\ref{sec:conclusion}.

\section{Related Work}\label{sec:related}

We summarize below the related work, which has treated a) temporally
i.i.d. rewards, b) rested Markovian rewards, and c) restless
Markovian rewards.

\subsection{Temporally i.i.d. rewards}


Lai and Robbins~\cite{Lai:Robbins} wrote one of the earliest papers
on the classic non-Bayesian infinite horizon multi-armed bandit
problem. They assume $K$ independent arms, each generating rewards
that are i.i.d. over time obtained from a distribution that can be
characterized by a single-parameter. For this problem, they present
a policy that provides an expected regret that is $O(K \log n)$,
i.e. linear in the number of arms and asymptotically logarithmic in
n. Anantharam \emph{et al.} extend this work to the case when $M$
simultaneous plays are allowed~\cite{Anantharam}. The work by
Agrawal~\cite{Agrawal:1995} presents easier to compute policies
based on the sample mean that also has asymptotically logarithmic
regret. The paper by Auer \emph{et al.}~\cite{Auer:2002} that
considers arms with nonnegative rewards that are i.i.d. over time
with an arbitrary non-parameterized distribution that has the only
restriction that it have a finite support. Further, they provide a
simple policy (referred to as UCB1), which achieves logarithmic
regret uniformly over time, rather than only asymptotically. Our
work utilizes a general Chernoff-Hoeffding-bound-based approach to
regret analysis pioneered by Auer \emph{et al.}.

Some recent work has shown the design of distributed multiuser
policies for independent arms. Motivated by the problem of
opportunistic access in cognitive radio networks, Liu and
Zhao~\cite{Liu:Zhao}, Anandkumar \emph{et
al.}~\cite{Anandkumar:Infocom:2010, Anandkumar:JSAC}, and Gai and
Krishnamachari~\cite{Gai:decentralized:globecom}, have developed
policies for the problem of $M$ distributed players operating $N$
independent arms.



Our work in this paper is closest to and builds on the recent work
by Gai \emph{et al.} which introduced combinatorial multi-armed
bandits~\cite{Gai:LLR}. The formulation in~\cite{Gai:LLR} has the
restriction that the reward process must be i.i.d. over time. A
polynomial storage learning algorithm is presented in~\cite{Gai:LLR}
that yields regret that is polynomial in users and resources and
uniformly logarithmic in time for the case of i.i.d. rewards.


\subsection{Rested Markovian rewards}

There has been relatively less work on multi-armed bandits with
Markovian rewards. Anantharam \emph{et al.}~\cite{Anantharam:1987}
wrote one of the earliest papers with such a setting. They proposed
a policy to pick $m$ out of the $N$ arms each time slot and prove
the lower bound and the upper bound on regret. However, the rewards
in their work are assumed to be generated by \emph{rested} (i.e.
rewards that only evolve when the arms are selected) Markov chains
with transition probability matrices defined by a single parameter
$\theta$ with identical state spaces. Also, for the upper bound the
result is achieved only asymptotically.

For the case of single users and independent arms, a recent work by
Tekin and Liu~\cite{Tekin:2010} has extended the results in
\cite{Anantharam:1987} relaxing the requirement of a single
parameter and identical state spaces across arms. They propose to
use UCB1 from \cite{Auer:2002} for the multi-armed bandit problem
with rested Markovian rewards and prove a logarithmic upper bound on
the regret under some conditions on the Markov chain.

In a recent work by Gai \emph{et al.}~\cite{Gai:rested:globecom},
learning policies for combinatorial multi-armed bandits with rested
Markovian rewards have been studied. Unlike
\cite{Gai:rested:globecom}, we adopt a model with restless Markovian
rewards, which has much broader applications in many network
optimization problems.

\subsection{Restless Markovian rewards}

Restless arm bandits are so named because the arms evolve at each
time, changing state even when they are not selected. Work on
restless Markovian rewards with single users and independent arms
can be found in \cite{Tekin:restless:infocom, Qing:icassp, Qing:ita,
Dai:icassp}. In these papers there is no consideration of possible
dependencies among arms, as in our work here.

Tekin and Liu \cite{Tekin:restless:infocom} have proposed a RCA
policy that achieves logarithmic single-action regret when certain
knowledge about the system is known. We use elements of the policy
and proof from~\cite{Tekin:restless:infocom} in this work, which is
however quite different in its combinatorial matching formulation
(which allows for dependent arms). Liu \emph{et
al.}~\cite{Qing:icassp, Qing:ita} adopted the same problem
formulation as in \cite{Tekin:restless:infocom}, and proposed a
policy named RUCB, achieving a logarithmic single-action regret over
time when certain system knowledge is known. They also extend the
RUCB policy to achieve a near-logarithmic regret asymptotically when
no knowledge about the system is available.

Dai \emph{et al.} in ~\cite{Dai:icassp} adopt a stronger definition
of regret: the difference in expected reward compared to a
model-aware genie. They develop a policy that yields regret of order
arbitrarily close to logarithmic for certain classes of restless
bandits with a finite-option structure, such as restless MAB with
two states and identical probability transition matrices.

\section{Problem Formulation}\label{sec:formulation}

We consider a system with $N$ edges predefined by some application,
where time is slotted and indexed by $n$. For each edge $i$ ($1 \leq
i \leq N$), there is an associated state that evolves as a
discrete-time, finite-state, aperiodic, irreducible Markov
chain\footnote{We also refer Markov chain $\{X^i(n), n \geq 0\}$ and
Markov chain $i$ interchangeably.} $\{X^i(n), n \geq 0\}$  with
unknown parameters\footnote{Alternatively, for Markov chain
$\{X^i(n), n \geq 0\}$, it suffices to assume that the
multiplicative symmetrization of the transition probability matrix
is irreducible.}. We denote the state space for the $i$-th Markov
chain by $S^i$. We assume these $N$ Markov chains are mutually
independent. The reward obtained from state $x$ ($x \in S^i$) of
Markov chain $i$ is denoted as $r^i_x$. Denote by $\pi^{i}_x$ the
steady state distribution for state $x$. The mean reward obtained on
Markov chain $i$ is denoted by $\mu^{i}$. Then we have $\mu^{i} =
\sum\limits_{z \in S_{i,j}} r^i_x \pi^i_x$. The set of all mean
rewards is denoted by $\bm{\mu} = \{\mu^{i}\}$.

At each decision period $n$ (also referred to interchangeably as
time slot),  an $N$-dimensional action vector $\xa(n)$, representing
an arm, is selected under a policy $\phi(n)$ from a finite set
$\xF$. We assume $a_i(n) \geq 0$ for all $1 \leq i \leq N$. When a
particular $\xa(n)$ is selected, the value of $r^{i}_{x_i(n)}$ is
observed, only for those $i$ with $a_i(n) \neq 0$. We denote by
$\xA_{\xa(n)} = \{i: a_i(n) \neq 0, 1 \leq i \leq N \}$ the index
set of all $a_i(n) \neq 0$ for an arm $\xa$. We treat each $\xa(n)
\in \xF$ as an arm. The reward is defined as:
\begin{equation}\label{equ:def}
  R^{\xa(n)}(n) = \sum\limits_{i \in \xA_{\xa(n)}} a_i(n) r^{i}_{x_i
  (n)}
\end{equation}
where $x_i(n)$ denotes the state of a Markov chain $i$ at time $n$.

When a particular arm $\xa(n)$ is selected, the rewards
corresponding to non-zero components of $\xa(n)$ are revealed,
i.e., the value of $r^{i}_{x_i(n)}$ is observed for all $i$ such
 that $a_i(n) \neq 0$.


The state of the Markov chain evolves \emph{restlessly}, i.e., the
state will continue to evolve independently of the actions. We
denote by $P^{i} = (p^{i}_{x,y})_{x, y \in S^{i}}$ the transition
probability matrix for the Markov chain $i$. We denote by $(P^i)' =
\{(p^i)'_{x,y}\}_{x, y \in S^{i}}$ the \emph{adjoint} of $P^i$ on
$l_2(\pi)$, so $(p^i)'_{x,y} = p^{i}_{y,x} \pi^i_y/\pi^i_x$. Denote
$\hat{P}^i=(P^i)'P$ as the \emph{multiplicative symmetrization} of
$P^i$. For aperiodic irreducible Markov chains, $\hat{P}^i$s are
irreducible \cite{Diaconis:1995}.


A key metric of interest in evaluating a given policy $\phi$ for
this problem is \emph{regret}, which is defined as the difference
between the expected reward that could be obtained by the
best-possible static action, and that obtained by the given policy.
It can be expressed as:
\begin{equation}
\begin{split}
 \mathfrak{R}^\phi (n) & = n \gamma^*  - \mathbb{E}^\phi[ \sum \limits_{t = 1}^n R^{\phi(t)}(t) ]\\
 & = n \gamma^*  - \mathbb{E}^\phi[ \sum \limits_{t = 1}^n \sum\limits_{i \in \xA_{\xa(t)}} a_i(t) r^{i}_{x_i(t)}]
 \end{split}
\end{equation}
where $\gamma^* = \max\limits_{\xa \in\xF} \sum\limits_{i \in
\xA_{\xa(n)}} a_i \mu^{i}$ is the expected reward of the optimal
arm. For the rest of the paper, we use $*$ as the index indicating
that a parameter is for an optimal arm. If there is more than one
optimal arm, $*$ refers to any one of them. We denote by
$\gamma^\xa$ the expected reward of arm $\xa$, so $\gamma^{\xa} =
\sum\limits_{j = 1}^{|\xA_{\xa}|} a_{p_j} \mu^{p_j}$.

For this combinatorial multi-armed bandit problem with restless
Markovian rewards, our goal is to design policies that perform well
with respect to regret. Intuitively, we would like the regret
$\mathfrak{R}^\phi (n)$ to be as small as possible. If it is
sublinear with respect to time $n$, the time-averaged regret will
tend to zero.

\section{Policy Design} \label{sec:algorithm:restless}

For the above combinatorial MAB problem with restless rewards, we
have two challenges here for the policy design:

(1) A straightforward idea is to apply RCA in
\cite{Tekin:restless:infocom}, or RUCB in \cite{Qing:icassp}
directly and naively, and ignore the dependencies across the
different arms. However, we note that RCA and RUCB both require the
storage and computation time that are linear in the number of arms.
Since there could be exponentially many arms in this formulation, it
is highly unsatisfactory.

(2) Unlike our prior work on combinatorial MAB with rested rewards,
for which the transitions only occur each time the Markov chains are
observed, the policy design for the restless case is much more
difficult, since the current state while starting to play a Markov
chain depends not only on the transition probabilities, but also on
the policy.


To deal with the first challenge, we want to design a policy which
more efficiently stores observations from the correlated arms, and
exploits the correlations to make better decisions. Instead of
storing the information for each arm, our idea is to use two $1$ by
$N$ vectors to store the information for each Markov chain. Then an
\emph{index} for each each arm is calculated, based on the
information stored for underlying components. This \emph{index} is
used for choosing the arm to be played each time when a decion needs
to be made.

To deal with the second challenge, for each arm $\xa$ we note that
the multidimensional Markov chain $\{X^{\xa}(n), n \geq 0\}$ defined
by underlying components as $X^{\xa}(n) = (X^{i}(n))_{i \in
\xA_{\xa}}$ is aperiodic and irreducible. Instead of utilizing the
actual sample path of all observations, we only take the
observations from a regenerative cycle for Markov chains and discard
the rest in its estimation of the \emph{index}.

Our proposed policy, which we refer to as Combinatorial Learning
with Restless Markov Reward (CLRMR), is shown in Algorithm
\ref{alg:restless}. Table \ref{tab:alg} summerizes the notation we
use for CLRMR algorithm. For Algorithm \ref{alg:restless}, $(x_i)_{i
\in \xA_{\xa}} = (\zeta^i)_{i \in \xA_{\xa}}$ means $x_i = \zeta^i,
\forall i$.

\begin{algorithm} [h!]
\caption{Combinatorial Learning with Restless Markov Reward (CLRMR)}
\label{alg:restless}

\begin{algorithmic}[1]
\State \commnt{ Initialization} \label{line:1}


\State $t=1$, $t_2=1$;

\State $\forall i=1,\cdots,N$, $m^i_2=0$, $\bar{z}^i_2=0$;

\For {$b = 1$ to $N$}
    \State $t := t+1$, $t_2 := t_2 + 1$;
    \State Play any arm $\xa$ such that $b \in \xA_\xa$; denote $(x_i)_{i \in \xA_{\xa}}$
    as the observed state vector for arm $\xa$;
    \State $\forall i \in \xA_{\xa(n)}$,
    let $\zeta^i$ be the first state observed for Markov chain $i$ if $\zeta^i$ has never been set;
    $\bar{z}^i_2 := \frac{\bar{z}^i_2 m^i_2 + r^i_{x_i}}{m^i_2 + 1}$, $m^i_2 := m^i_2 + 1$;
    \While {$(x_i)_{i \in \xA_{\xa}} \neq (\zeta^i)_{i \in \xA_{\xa}}$}
        \State $t := t+1$, $t_2 := t_2 + 1$;
        \State Play arm $\xa$; denote $(x_i)_{i \in \xA_{\xa}}$
    as the observed state vector;
        \State $\forall i \in \xA_{\xa(n)}$, $\bar{z}^i_2 := \frac{\bar{z}^i_2 m^i_2 + r^i_{x_i}}{m^i_2 + 1}$, $m^i_2 := m^i_2 + 1$;
    \EndWhile
\EndFor  \label{line:13}

\State \commnt{Main loop}

\While {1}
    \State \commnt{SB1 starts}
    \State $t := t+1$;
    \State Play an arm $\xa$ which maximizes
    \begin{equation}
    \label{equ:maxrestless}
    \max\limits_{\xa \in \xF} \sum\limits_{i \in \xA_{\xa}} a_i \left(\bar{z}^i_2 + \sqrt{ \frac{ L \ln t_2 }{ m^i_2}} \right);
    \end{equation}
    \State Denote $(x_i)_{i \in \xA_{\xa}}$ as the observed state vector;
    \While {$(x_i)_{i \in \xA_{\xa}} \neq (\zeta^i)_{i \in \xA_{\xa}}$}
        \State $t := t+1$;
        \State Play an arm $\xa$ and denote $(x_i)_{i \in \xA_{\xa}}$ as the observed state
        vector;
    \EndWhile
    \State \commnt{SB2 starts}
    \State $t_2 := t_2 + 1$;
    \State $\forall i \in \xA_{\xa(n)}$, $\bar{z}^i_2 := \frac{\bar{z}^i_2 m^i_2 + r^i_{x_i}}{m^i_2 + 1}$, $m^i_2 := m^i_2 + 1$;
    \While {$(x_i)_{i \in \xA_{\xa}} \neq (\zeta^i)_{i \in \xA_{\xa}}$}
        \State $t := t+1$, $t_2 := t_2 + 1$;
        \State Play an arm $\xa$ and denote $(x_i)_{i \in \xA_{\xa}}$ as the observed state
        vector;
        \State $\forall i \in \xA_{\xa(n)}$, $\bar{z}^i_2 := \frac{\bar{z}^i_2 m^i_2 + r^i_{x_i}}{m^i_2 + 1}$, $m^i_2 := m^i_2 + 1$;
    \EndWhile
    \State \commnt{SB3 is the last play in the while loop. Then a block completes.}
    \State $b: = b + 1$, $t := t+1$;
\EndWhile
\end{algorithmic}
\end{algorithm}

\begin{table}[htbp]
\centering \normalsize
\begin{tabular}{|l@{\hspace{0.6mm}}l@{}|}
\hline
 $N$ : & number of resources\\
 $\xa$: & vectors of coefficients, defined on set $\xF$; \\
  & we map each $\xa$ as an arm\\
 $\xA_\xa$: & $\{i: a_i \neq 0, 1 \leq i \leq N \}$\\
 $t$: & current time slot\\
 $t_2$: & number of time slots in SB2 up to the current \\
    & time slot\\
 $b$: & number of blocks up to the current time slot\\
 $m^i_2$: & number of times that Markov chain $i$ has been  \\
  & observed during SB2 up to the current time slot\\
 $\bar{z}^i_2$: & average (sample mean) of all the observed \\
  & values of Markov chain $i$ during SB2 up to \\
  & the current time slot\\
 $\zeta^i$: & state that determine the regenerative cycles for \\
   & Markov chain $i$\\
 $x_i$: & the observed state when Markov Chain $i$ is \\
   & played; $(x_i)_{i \in \xA_{\xa}}$ is the observed state vector \\
   & if arm $\xa$ is played\\
\hline
\end{tabular}
\caption{Notation for Algorithm \ref{alg:restless}} \label{tab:alg}
\end{table}

\begin{figure}[t]
\vspace{0.08in}
\includegraphics[width=0.48\textwidth]{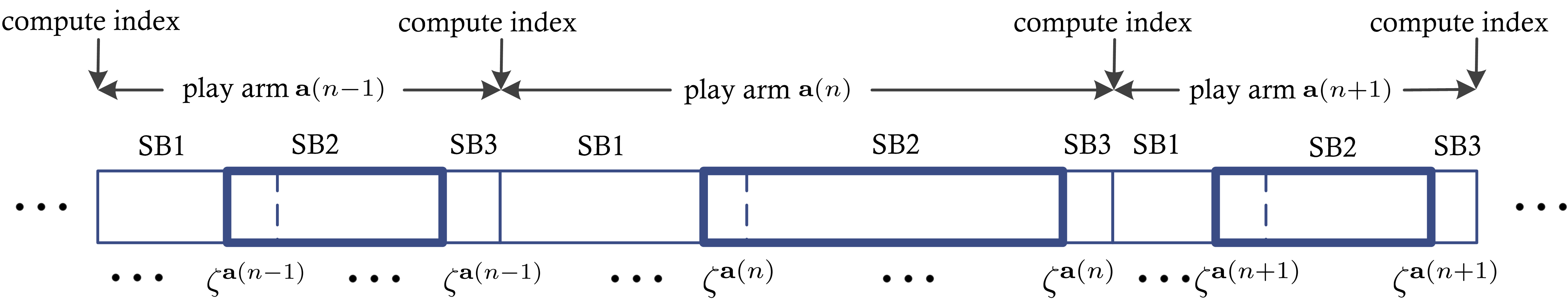}
\caption{An illustration of CLRMR} \label{fig:1}
\end{figure}

CLRMR operates in blocks. Figure \ref{fig:1} illustrates one
possible realization of this Algorithm \ref{alg:restless}. At the
beginning of each block, an arm $\xa$ is picked and within one
block, this algorithm always play the same arm. For each Markov
chain $\{X^{i}(n)\}$, we specifiy a state $\zeta^i$ at the beginning
of the algorithm as a state to mark the regenerative cycle. Then,
for the multidimentional Markov chain $\{X^{\xa}(n)\}$ associated
with this arm, the state $(\zeta^i)_{i \in \xA_{\xa}}$ is used to
define a regenerative cycle for $\{X^{\xa}(n)\}$.

Each block is broken into three sub-blocks denoted by SB1, SB2 and
SB3. In SB1, the selected arm $\xa$ is played until the state
$(\zeta^i)_{i \in \xA_{\xa}}$ is observed. Upon this observation we
enter a regenerative cycle, and continue playing the same arm untill
$(\zeta^i)_{i \in \xA_{\xa}}$ is observed again. SB2 includes all
time slots from the first visit of $(\zeta^i)_{i \in \xA_{\xa}}$ up
to but excluding the second visit to $(\zeta^i)_{i \in \xA_{\xa}}$.
SB3 consists a single time slot with the second visit to
$(\zeta^i)_{i \in \xA_{\xa}}$. SB1 is empty if the first observed
state is $(\zeta^i)_{i \in \xA_{\xa}}$. So SB2 includes the observed
rewards for a regenerative cycle of the multidimentional Markov
chain $\{X^{\xa}(n)\}$ associated with arm $\xa$, which implies that
SB2 also includes the observed rewards for one or more regenerative
cycles for each underlying Markov chain $\{X^i(n)\}, i \in
\xA_{\xa}$.

The key to the algorithm \ref{alg:restless} is to store the
observations for each Markov chain instead of the whole arm, and
utilize the observations only in SB2 for them, and virtually
assemble them (highlighted with thick lines in Figure \ref{fig:1}).
Due to the regenerative nature of the Markov chain, by putting the
observations in SB2, the sample path has exactly the same statics as
given by the transition probability matrix. So the problem is
tractable.

LLR policy requires storage linear in $N$. We use two $1$ by $N$
vectors to store the information for each Markov chain after we play
the selected arm at each time slot in SB2. One is $(\bar{z}^i_2)_{1
\times N}$ in which $\bar{z}^i_2$ is the average (sample mean) of
observed values in SB2 up to the current time slot (obtained through
potentially different sets of arms over time). The other one is
$(m^i_2)_{1 \times N}$ in which $m^i_2$ is the number of times that
$\{X^i(n)\}$ has been observed in SB2 up to the current time slot.

Line \ref{line:1} to line \ref{line:13} are the initialization, for
which each Markov chain is observed at least once, and $\zeta^i$ is
specified as the first state observed for $\{X^i(n)\}$.

After the initialization, at the beginning of each block, CLRMR
selects the arm which solves the maximization problem as in
(\ref{equ:maxrestless}). It is a deterministic linear optimal
problem with a feasible set $\xF$ and the computation time for an
arbitrary $\xF$ may not be polynomial in $N$. But, as we show in
Section \ref{sec:app:simulation}, there exist many practically
useful examples with polynomial computation time.

\section{Analysis of Regret} \label{sec:regret}

We summarize some notation we use in the description and analysis of
our CLRMR policy in Table \ref{tab:notation:restless}.

\begin{table}[htbp]
\centering \normalsize
\begin{tabular}{|l l|}
\hline
 $H$ : & $\max\limits_\xa |\xA_\xa|$. Note that $H \leq N$\\
 $\xa(\tau)$ : & the arm played in time $\tau$\\
 $b(n)$: & number of completed blocks up to time $n$ \\
 $t(b)$: & time at the end of block $b$\\
 $t_2(b)$: & total number of time slots spent in SB2  \\
  & up to block $b$\\
 $B^{\xa}(b)$: & total number of blocks within the first $b$ \\
    & blocks in which arm $\xa$ is played \\
 \multicolumn{2}{|l|}{$m^i_2(t_2(b))$:  total number of time slots Markov chain $i$} \\
  & is observed during SB2 up to block $b$ \\
 $\bar{z}^i_2(s)$: & the mean reward from Markov chain $i$ \\
  & when it is observed for the $s$-th time of \\
  & only those times played during SB2\\
 $T(n)$: & time at the end of the last completed block\\
 $T^{\xa}(n)$: & total number of time slots arm $\xa$ is palyed \\
  & up to time $T(n)$\\
 $m^i_x(s)$: & number of times that state $x$ occured when \\
  & Markov chain $i$
 has been observed $s$ times\\
 $Y^i_1(j)$: & vector of observed states from SB1 of the \\
  & $j$-th block for playing Markov chain $i$ \\
 $Y^i_2(j)$: & vector of observed states from SB2 of the \\
   & $j$-th block for playing Markov chain $i$ \\
 $Y^i(j)$: & vector of observed states from the $j$-th \\
  & block for playing Markov chain $i$ \\
 $\hat{\pi}^{i}_x$: & $\max\{\pi^i_x, 1 - \pi^i_x\}$\\
 $\hat{\pi}_{\max}$: & $\max\limits_{i, x \in S^i}\hat{\pi}^{i}_x$\\
 $\pi_{\min}$: & $\min\limits_{i, x \in S^i} \pi^i_x$\\
 $\pi_{\max}$: & $\max\limits_{i, x \in S^i} \pi^i_x$\\
 $\epsilon^{i}$: & eigenvalue gap, defined as $1 - \lambda_2$, where  \\
 &  $\lambda_2$ is the second largest eigenvalue of the \\
 & multiplicative symmetrization of $P^i$\\
 $\epsilon_{\min}$: & $\min\limits_{i} \epsilon^i$ \\
 $S_{\max}$: & $\max\limits_{i} |S^i|$\\
 $r_{\max}$: & $\max\limits_{i, x \in S^i}r^{i}_x$\\
 $a_{\max}$: & $\max\limits_{i \in \xA_\xa, \xa \in \xF} a_i$\\
 $\Delta_\xa$: & $\gamma^* - \gamma^{\xa}$\\
 $\Delta_{\min}$: & $\min\limits_{\gamma^{\xa} \leq \gamma^*} \Delta_\xa$\\
 $\Delta_{\max}$: & $\max\limits_{\gamma^{\xa} \leq \gamma^*} \Delta_\xa$\\
 \multicolumn{2}{|l|}{$\{X^{\xa}(n)\}$: multidimentional Markov chain defined} \\
   & by $X^{\xa}(n) = (X^{i}(n))_{i \in \xA_{\xa}}$\\
 $\zeta^{\xa}$: & $(\zeta^{i})_{i \in \xA_{\xa}}$, state vector that determines \\
  & the regenerative cycles for $\{X^{\xa}(n)\}$ \\
 $\Pi^{\xa}_z$: & steady state distribution for state $z$ of $\{X^{\xa}(n)\}$\\
 $\Pi^{\xa}_{\min}$: & $\min\limits_{z \in S^{\xa}} \Pi^{\xa}_z$\\
 $\Pi_{\min}$: & $\min\limits_{\xa, z \in S^{\xa}} \Pi^{\xa}_z$ \\
 $M^{\xa}_{z_1, z_2}$: & mean hitting time of state $z_2$ starting \\
   & from an initial state $z_1$ for $\{X^{\xa}(n)\}$\\
 $M^{\xa}_{\max}$: & $\max\limits_{z_1, z_2 \in S^{\xa}} M^{\xa}_{z_1, z_2}$ \\
 $\gamma'_{\max}$: & $\max\limits_{\gamma^{\xa} \leq \gamma^*} \gamma^{\xa}$\\
\hline
\end{tabular}
\caption{Notation for Regret Analysis} \label{tab:notation:restless}
\end{table}


We first show in Theorem \ref{theorem:1} an upper bound on the total
expected number of plays of suboptimal arms.

\begin{theorem}\label{theorem:1}
When using any constant $L \geq 56(H+1) S^2_{\max} r^{2}_{\max}
\hat{\pi}^{2}_{\max}/\epsilon_{\min}$, we have
\begin{align}
 & \sum\limits_{\xa: \gamma^{\xa} < \gamma^*} (\gamma^* -
\gamma^{\xa}) \mathbb{E}[T^{\xa}(n)]  \leq Z_1 \ln n + Z_2 \nonumber
\end{align}
where
\begin{align}
& Z_1 = \Delta_{\max} \left(\frac{1}{\Pi_{\min}} + M_{\max} +1
\right) \frac{4 N L H^2 a_{\max}^2}{ \Delta^2_{\min} } \nonumber\\
& Z_2 =  \Delta_{\max} \left(\frac{1}{\Pi_{\min}} + M_{\max} +1
\right) \left(N + \frac{\pi N H S_{\max}}{3 \pi_{\min} } \right)
\nonumber
\end{align}
\end{theorem}

To proof Theorem \ref{theorem:1}, we use the inequalities as stated
in Theorem 3.3 from \cite{lezaud} and a theorem from \cite{bremaud}.

\lemma[Theorem 3.3 from \cite{lezaud}] \label{lemma:1} Consider a
finite-state, irreducible Markov chain $\left\{X_t\right\}_{t \geq
1}$ with state space $S$, matrix of transition probabilities $P$, an
initial distribution $\mathbf{q}$ and stationary distribution
$\mathbf{\pi}$. Let $N_{\mathbf{q}}=\left\|(\frac{q_x}{\pi_x}, x\in
S)\right\|_2$. Let $\hat{P}=P'P$ be the multiplicative
symmetrization of $P$ where $P'$ is the adjoint of $P$ on
$l_2(\pi)$. Let $\epsilon=1-\lambda_2$, where $\lambda_2$ is the
second largest eigenvalue of the matrix $\hat{P}$. $\epsilon$ will
be referred to as the eigenvalue gap of $\hat{P}$. Let
$f:S\rightarrow \mathcal{R}$ be such that $\sum_{y \in S} \pi_y f(y)
=0$, $\left\|f\right\|_{\infty} \leq 1$ and $0<\left\|f\right\|^2_2
\leq 1$. If $\hat{P}$ is irreducible, then for any positive integer
$n$ and all $0 < \delta \leq 1$
\begin{eqnarray*}
P\left(\frac{\sum_{t=1}^n f(X_t)}{n} \geq \delta \right) \leq N_q
\exp\left[-\frac{n \delta^2 \epsilon}{28}\right] ~.
\end{eqnarray*}

\lemma \label{lemma:2} If $\left\{X_n\right\}_{n \geq 0}$ is a
positive recurrent homogeneous Markov chain with state space $S$,
stationary distribution $\pi$ and $\tau$ is a stopping time that is
finite almost surely for which $X_{\tau}=x$ then for all $y \in S$
\begin{eqnarray*}
E\left[ \sum_{t=0}^{\tau-1} I(X_t=y) | X_0=x \right] = E[\tau |
X_0=x]\pi_y ~.
\end{eqnarray*}

\begin{IEEEproof} [Proof of Theorem \ref{theorem:1}]



We introduce $\xB(b)$ as a counter for the regret analysis to deal
with the combinatorial arms. After the initialization period,
$\xB(b)$ is updated in the following way: at the beginning of any
block when a non-optimal arm is chosen to be played, find $i$ such
that $i =  \arg \min\limits_{ j \in \xA_\xa(b) } m_2^{j}$ ($i$ the
index of the elements which are among the ones that have been
observed least in SB2 in the non-optimal arm). If there is only one
such arm, $\xB(b)$ is increased by $1$. If there are multiple such
arms, we arbitrarily pick one, say $i'$, and increment
$\widetilde{B}^{i'}$ by $1$. Based on the above definition of
$\xB(b)$, each time a non-optimal arm is chosen to be played at the
beginning of a block, exactly one element in $(\xB(b))_{1 \times N}$
is incremented by $1$. So the summation of all counters in
$(\xB(b))_{1 \times N}$ equals the total number of blocks in which
we have played non-optimal arms,
\begin{equation}
 \label{equ:f1}
 \sum\limits_{\xa: \gamma^{\xa} < \gamma^*} \mathbb{E}[B^\xa(b)] = \sum\limits_{i =
 1}^{N}
 \mathbb{E}[\xB(b)].
\end{equation}

We also have the following inequality for $\xB(b)$:
\begin{equation}
 \label{equ:f2}
 \xB(b) \leq m^{i}_2(t(b-1)), \forall 1 \leq i \leq N, \forall b.
\end{equation}

Denote by $c_{t, s}$ $\sqrt{ \frac{ L \ln t }{ s} }$. Denote by
$\uI(b)$ the indicator function which is equal to $1$ if $\xB(b)$ is
added by one at block $b$. Let $l$ be an arbitrary positive integer.
Then we can get the upper bound of $\mathds{E}[\xB(b)]$ shown in
(\ref{equ:14}),

\begin{equation}
\begin{split}
 & \mathds{E}[\xB(b)]  = \sum\limits_{\beta = N+1}^b \mathds{P} \{
 \uI(\beta)=1\}\\
 & \leq l + \sum\limits_{\beta = N+1}^b \mathds{P} \{ \uI(\beta)=1 , \xB(\beta-1) \geq l \} \\
 & \leq l + \sum\limits_{\beta = N+1}^b \mathds{P}\{ \sum\limits_{k \in \xA_{\xa^*}}
 a_k^* g^k_{t_2(\beta-1), m^k_2(t(\beta-1))}\\
 & \quad \leq  \sum\limits_{j \in \xA_{\xa(h)}} a_j(b) g^{j}_{t_2(\beta-1), m^j_2(t(\beta-1))}
 , \xB(\beta-1) \geq l \}.  \label{equ:14}
\end{split}
\end{equation}
where $g^i_{t,s}=\xZ^{i}(s)+c_{t,s}$ and $\xa(\beta)$ is defined as
a non-optimal arm picked at block $\beta$ when $\uI(\beta) = 1$.
Note that $m^{i}_2 = \min\limits_j \{m^j_2: \forall j \in
\xA_{\xa(\beta)}\}$. We denote this arm by $\xa(\beta)$ since at
each block that $\uI(\beta) = 1$, we could get different arms.

Note that $l \leq \xB(\beta-1)$ implies,
\begin{equation} \label{equ:13}
 l \leq \xB(\beta-1) \leq m^{i}_2(t(\beta-1)), \forall j \in \xA_{\xa(\beta)}.
\end{equation}

So we can further derive the upper bound of $\mathds{E}[\xB(b)]$
shown in (\ref{equ:15}), where $h_j$ ($1 \leq j \leq |\xA_{\xa*}|$)
represents the $j$-th element in $\xA_{\xa*}$; $p_j$ ($1 \leq j \leq
|\xA_{\xa(\beta)}|$) represents the $j$-th element in
$\xA_{\xa(\beta)}$ or $\xA_{\xa(t)}$. $\xA_{\xa(\tau)}$ represents
the arm played in the $\tau$-th time slots counting only in SB2.
Note that
\begin{figure*}[t]
\normalsize \setcounter{equation}{14}
\begin{align}
 \mathds{E}[\xB(b)] & \leq l + \sum\limits_{\beta = N+1}^b
 \mathds{P} \{ \min\limits_{0 < s_{h_1}, \ldots, s_{h_{|\xA_{\xa*}|}} < t_2(\beta-1) }   \sum\limits_{j = 1}^{|\xA_{\xa*}|}
     a_{h_j}^* g^{h_j}_{t_2(\beta-1), s_{h_j}}
  \leq \max\limits_{t_2(l) \leq s_{p_1}, \ldots, s_{p_{|\xA_{\xa(\beta)}|}} < t_2(\beta-1)} \sum\limits_{j = 1}^{|\xA_{\xa(\beta)}|}
    a_{p_j}(\beta) g^{p_j}_{t_2(\beta-1), s_{p_j}} \} \nonumber\\
 & \leq l + \sum\limits_{\beta = N+1}^{b} \sum\limits_{s_{h_1} = 1}^{t_2(\beta-1)} \dots \sum\limits_{s_{h_{|\xA^*|}} = 1}^{t_2(\beta-1)}
 \sum\limits_{s_{p_1} = t_2(l)}^{t_2(\beta-1)} \dots \sum\limits_{s_{p_{|\xA_{\xa(\beta)}|}} = t_2(l)}^{t_2(\beta-1)}
    \mathds{P}\{\sum\limits_{j = 1}^{|\xA_{\xa*}|} a_{h_j}^* g^{h_j}_{t_2(\beta-1), s_{h_j}}
  \leq \sum\limits_{j = 1}^{|\xA_{\xa(\beta)}|} a_{p_j}(\beta) g^{p_j}_{t_2(\beta-1), s_{p_j}}
 \} \nonumber \\
  & \leq l + \sum\limits_{\tau = 1}^{t_2(b)} \sum\limits_{s_{h_1} = 1}^{\tau-1} \dots \sum\limits_{s_{h_{|\xA^*|}} = 1}^{\tau-1}
 \sum\limits_{s_{p_1} = l}^{\tau-1} \dots \sum\limits_{s_{p_{|\xA_{\xa(\beta)}|}} = l}^{\tau-1}
    \mathds{P}\{\sum\limits_{j = 1}^{|\xA_{\xa*}|} a_{h_j}^* g^{h_j}_{\tau, s_{h_j}}
  \leq \sum\limits_{j = 1}^{|\xA_{\xa(\tau)}|} a_{p_j}(\tau) g^{p_j}_{\tau, s_{p_j}}
 \} \label{equ:15}
\end{align}
\setcounter{equation}{15} \hrulefill \vspace*{4pt}
\end{figure*}

\begin{align}
  & \mathds{P}\{\sum\limits_{j = 1}^{|\xA_{\xa*}|} a_{h_j}^* g^{h_j}_{\tau, s_{h_j}}
  \leq \sum\limits_{j = 1}^{|\xA_{\xa(\tau)}|} a_{p_j}(t) g^{p_j}_{\tau, s_{p_j}}
 \}\\
  & = \mathds{P}\{\sum\limits_{j = 1}^{|\xA_{\xa*}|} a_{h_j}^* ( \xZ^{h_j}(s_{h_j})+c_{\tau,s_{h_j}}
    ) \\
  & \qquad \leq \sum\limits_{j = 1}^{|\xA_{\xa(\tau)}|} a_{p_j}(\tau) ( \xZ^{p_j}(s_{p_j})+c_{\tau,s_{p_j}} )
 \} \label{equ:15a} \\
 &  = \mathds{P}\{ \text{At least one of the following must hold:} \nonumber\\
  &  \sum\limits_{j = 1}^{|\xA_{\xa*}|} a_{h_j}^* \xZ^{h_j}(s_{h_j}) \leq  \gamma^* - \sum\limits_{j =
  1}^{|\xA_{\xa*}|} a_{h_j}^* c_{\tau, s_{h_j}},\label{equ:ineq1}\\
  &  \sum\limits_{j = 1}^{|\xA_{\xa(\tau)}|} a_{p_j}(\tau) \xZ^{p_j}(s_{p_j})
 \geq \gamma^{\xa(\tau)} + \sum\limits_{j = 1}^{|\xA_{\xa(\tau)}|} a_{p_j}(\tau) c_{\tau,s_{p_j}} , \label{equ:ineq2}\\
  &  \gamma^* < \gamma^{\xa(\tau)}  + 2 \sum\limits_{j = 1}^{|\xA_{\xa(\tau)}|}  a_{p_j}(\tau) c_{\tau,s_{p_j}} \label{equ:ineq3}\}
\end{align}
Now we show the upper bound on the probabilities of inequalities
(\ref{equ:ineq1}), (\ref{equ:ineq2}) and (\ref{equ:ineq3})
separately. We first find an upper bound on the probability of
(\ref{equ:ineq1}):

\begin{align}
&\mathds{P}\{ \sum\limits_{j = 1}^{|\xA_{\xa*}|} a_{h_j}^*
\xZ^{h_j}(s_{h_j}) \leq  \gamma^* - \sum\limits_{j =
  1}^{|\xA_{\xa*}|} a_{h_j}^* c_{\tau, s_{h_j}}\} \nonumber\\
& = \mathds{P}\{ \sum\limits_{j = 1}^{|\xA_{\xa*}|} a_{h_j}^*
\xZ^{h_j}(s_{h_j}) \leq  \sum\limits_{j = 1}^{|\xA_{\xa*}|}
a_{h_j}^* \mu^{h_j} - \sum\limits_{j =
  1}^{|\xA_{\xa*}|} a_{h_j}^* c_{\tau, s_{h_j}}\} \nonumber\\
 & \leq \sum\limits_{j = 1}^{|\xA_{\xa*}|} \mathds{P}\{ a_{h_j}^* \xZ^{h_j}(s_{h_j}) \leq a_{h_j}^* (\mu^{h_j} - c_{\tau,
 s_{h_j}} ) \}\nonumber\\
 & = \sum\limits_{j = 1}^{|\xA_{\xa*}|} \mathds{P}\{\xZ^{h_j}(s_{h_j}) \leq \mu^{h_j} - c_{\tau,
 s_{h_j}}  \}. \nonumber
\end{align}

$\forall 1 \leq j \leq |\xA_{\xa*}|$,
\begin{align}
 & \mathds{P}\{\xZ^{h_j}(s_{h_j}) \leq \mu^{h_j} - c_{\tau,
 s_{h_j}}  \} \nonumber\\
 & = \mathds{P}\{ \sum\limits_{x \in S^{h_j}} ( \frac{r^{h_j}_x m^{h_j}_x(s_{h_j})}{s_{h_j}} - r^{h_j}_x \pi^{h_j}_x) \leq \sum\limits_{x \in S^{h_j}} -\frac{c_{\tau,
 s_{h_j}} }{|S^{h_j}|} \} \nonumber\\
 & \leq \sum\limits_{x \in S^{h_j}} \mathds{P}\{ \frac{r^{h_j}_x m^{h_j}_x(s_{h_j})}{s_{h_j}} - r^{h_j}_x \pi^{h_j}_x \leq -\frac{c_{\tau,
 s_{h_j}} }{|S^{h_j}|}  \} \nonumber\\
 & =  \sum\limits_{x \in S^{h_j}} \mathds{P}\{ r^{h_j}_x
 m^{h_j}_x(s_{h_j})  - s_{h_j} r^{h_j}_x \pi^{h_j}_x \leq -\frac{s_{h_j} c_{\tau,
 s_{h_j}} }{|S^{h_j}|} \} \nonumber
  \end{align}
 \begin{align}
 & = \sum\limits_{x \in S^{h_j}} \mathds{P}\{ r^{h_j}_x (s_{h_j} - \sum\limits_{y \neq x}
 m^{h_j}_y(s_{h_j}))\nonumber \\
 & \qquad\qquad - r^{h_j}_x s_{h_j} (1 - \sum\limits_{y \neq x}  \pi^{h_j}_y )\leq -\frac{s_{h_j} c_{\tau,
 s_{h_j}} }{|S^{h_j}|} \} \nonumber\\
 & = \sum\limits_{x \in S^{h_j}} \mathds{P}\{  \sum\limits_{y \neq
 x} m^{h_j}_y(s_{h_j}) -  \sum\limits_{y \neq x}
 \pi^{h_j}_y \geq \frac{s_{h_j} c_{\tau,
 s_{h_j}} }{r^{h_j}_x |S^{h_j}|}\nonumber\\
 & = \sum\limits_{x \in S^{h_j}} \mathds{P}\{ \frac{ \sum\limits_{t = 1}^{s_{h_j}}
 \mathds{1}(Y_{t}^{h_j} \neq x) - s_{h_j} (1-\pi^{h_j}_x)  }{\hat{\pi}^{h_j}_x s_{h_j}}\geq \frac{s_{h_j} c_{\tau,
 s_{h_j}} }{r^{h_j}_x |S^{h_j}|} \} \nonumber\\
 & \leq \sum\limits_{x \in S^{h_j}} N_{\mathbf{q}^{h_j}} \tau^{-\frac{L \epsilon^{h_j}}{28 (|S^{h_j}|r^{h_j}_x \hat{\pi}^{h_j}_x)^2}} \label{equ:18} \\
 & \leq \frac{|S^{h_j}|}{\pi_{\min}} \tau^{-\frac{L \epsilon_{\min}}{28 S^2_{\max} r^{2}_{\max}
 \hat{\pi}^{2}_{\max}}} \label{equ:19}
\end{align}
where (\ref{equ:18}) follows from Lemma \ref{lemma:1} by letting
\begin{align}
   \delta  = \frac{s_{h_j} c_{\tau,
 s_{h_j}} }{r^{h_j}_x |S^{h_j}|}, \;
  f(Y^{i}_t)  = \frac{
 \mathds{1}(Y_{t}^{i} \neq x) - (1-\pi^{i}_x)  }{\hat{\pi}^{i}_x }.
 \nonumber
\end{align}
$\mathds{1}(a)$ is the indicator function defined to be 1 when the
predicate $a$ is true, and 0 when it is false. $\hat{\pi}^{i}_x$ is
defined as $\hat{\pi}^{i}_x = \max\{\pi^i_x, 1-  \pi^i_x\}$ to
guarantee $\left\|f\right\|_{\infty} \leq 1$. We note that when
$\delta
> 1$ the deviation probability is zero, so the bound still holds.

(\ref{equ:19}) follows from the fact that for any $\mathbf{q}^{i}$,
\begin{equation}
\begin{split}
 N_{\mathbf{q}^{i}} & = \left\|\frac{q^{i}_x}{\pi^{i}_x}, x \in S^{i} \right\|_2 \leq \sum\limits_{x = 1}^{|S^{i}|}\left\|\frac{q^{i}_x}{\pi^{i}_x}
 \right\|_2 \leq \sum\limits_{x = 1}^{|S^{i}|}\frac{\left\|
 q^{i}_x\right\|_2}{\pi_{\min}} = \frac{1}{\pi_{\min}}. \nonumber
\end{split}
\end{equation}

Note that all the quantities in computing the indices and the
probabilities above come from SB2. Got for every SB2 in a block, the
quantities begin with state $\zeta^{\xa}$ and end with a return to
$\zeta^{\xa}$. So for each underlying Markov chain $\{X^i(n)\}, i
\in \xA_{\xa}$, the quantities got begin with state $\zeta^{i}$ and
end with a return to $\zeta^{i}$. Note that for all $i$, Markov
chain $\{X^i(n)\}$ could be played in different arms, but the
quantities got always begin with state $\zeta^{i}$ and end with a
return to $\zeta^{i}$. Then by the strong Markov property, the
process at these stopping times has the same distribution as the
original process. Connecting these intervals together we form a
continuous sample path which can be viewed as a sample path
generated by a Markov chain with transition matrix identical to the
original arm. This is the reason why we can apply Lemma
\ref{lemma:1} to this Markov chain.

Therefore,
\begin{align}
 &\mathds{P}\{ \sum\limits_{j = 1}^{|\xA_{\xa*}|} a_{h_j}^*
\xZ^{h_j}(s_{h_j}) \leq  \gamma^* - \sum\limits_{j =
  1}^{|\xA_{\xa*}|} a_{h_j}^* c_{\tau, s_{h_j}}\} \nonumber\\
  & \leq \frac{H S_{\max}}{\pi_{\min}} \tau^{-\frac{L \epsilon_{\min}}{28 S^2_{\max} r^{2}_{\max}
 \hat{\pi}^{2}_{\max}}} \label{equ:21}
\end{align}

With a similar derivation, we have
\begin{align}
& \mathds{P}\{ \sum\limits_{j = 1}^{|\xA_{\xa(\tau)}|} a_{p_j}(\tau)
\xZ^{p_j}(s_{p_j})
 \geq \gamma^{\xa(\tau)} + \sum\limits_{j = 1}^{|\xA_{\xa(\tau)}|} a_{p_j}(\tau)
 c_{\tau,s_{p_j}}\} \nonumber \\
 & \leq \sum\limits_{j = 1}^{|\xA_{\xa(\tau)}|} \mathds{P}\{ a_{p_j}(\tau) \xZ^{p_j}(s_{p_j})
  \geq a_{p_j}(\tau) \mu^{p_j} + a_{p_j}(\tau)
 c_{\tau,s_{p_j}} \} \nonumber \\
 & \leq \sum\limits_{j = 1}^{|\xA_{\xa(\tau)}|} \sum\limits_{x \in S^{p_j}} \mathds{P}\{ r^{p_j}_x
 m^{p_j}_x(s_{p_j})  - s_{p_j} r^{p_j}_x \pi^{p_j}_x \geq \frac{s_{p_j} c_{\tau,
 s_{p_j}} }{|S^{p_j}|} \} \nonumber\\
 & = \sum\limits_{j = 1}^{|\xA_{\xa(\tau)}|} \sum\limits_{x \in S^{p_j}} \mathds{P}\{ \frac{ \sum\limits_{t = 1}^{s_{p_j}}
 \mathds{1}(Y_{t}^{p_j} = x) - s_{p_j} \pi^{p_j}_x  }{\hat{\pi}^{p_j}_x s_{p_j}}\geq \frac{s_{p_j} c_{\tau,
 s_{p_j}} }{r^{p_j}_x |S^{p_j}|} \} \nonumber\\
  & \leq \sum\limits_{j = 1}^{|\xA_{\xa(\tau)}|} \sum\limits_{x \in S^{p_j}} N_{\mathbf{q}^{p_j}} \tau^{-\frac{L \epsilon^{p_j}}{28 (|S^{p_j}|r^{p_j}_x \hat{\pi}^{p_j}_x)^2}} \label{equ:22} \\
 & \leq \frac{H S_{\max}}{\pi_{\min}} \tau^{-\frac{L \epsilon_{\min}}{28 S^2_{\max} r^{2}_{\max}
 \hat{\pi}^{2}_{\max}}} \label{equ:23}
\end{align}
where (\ref{equ:22}) follows from Lemma \ref{lemma:1} by letting
\begin{align}
   \delta  = \frac{s_{p_j} c_{\tau,
 s_{p_j}} }{r^{p_j}_x |S^{p_j}|}, \;  f(Y^{i}_t)  = \frac{
 \mathds{1}(Y_{t}^{i} = x) - \pi^{i}_x  }{\hat{\pi}^{i}_x }.
 \nonumber
\end{align}

Note that when $l \geq \left\lceil \frac{4 L \ln t_2(b)}{
\left(\frac{\Delta_{\xa(\tau)}}{H a_{\max}} \right)^2 }
\right\rceil$, (\ref{equ:ineq3}) is false for $\tau$, which gives,

\begin{align}
 & \gamma^* - \gamma^{\xa(\tau)}  - 2 \sum\limits_{j = 1}^{|\xA_{\xa(\tau)}|} a_{p_j}(\tau) c_{\tau, s_{p_j}}  \nonumber\\
 & = \gamma^* - \gamma^{\xa(\tau)} - 2 \sum\limits_{j = 1}^{|\xA_{\xa(\tau)}|} a_{p_j} \sqrt{ \frac{ L \ln t_2(b) }{ s_{p_j}} } \nonumber\\
 & \geq \gamma^* - \gamma^{\xa(\tau)} - H a_{\max} \sqrt{ \frac{ 4L \ln t_2(b) }{ l } } \nonumber\\
 & \geq \gamma^* - \gamma^{\xa(\tau)} - H a_{\max} \sqrt{ \frac{ 4L \ln t_2(b) }{ 4L \ln t_2(b)} \left( \frac{\Delta_{\xa(t)} }{H a_{\max}} \right)^2 } \\
 & \geq \gamma^* - \gamma^{\xa(\tau)} - \Delta_{\xa(\tau)} = 0.
 \label{equ:25}
\end{align}

Hence, when we let $l \geq \left\lceil \frac{4 L H^2 a_{\max}^2 \ln
t_2(b)}{ \Delta^2_{\min} } \right\rceil$, (\ref{equ:ineq3}) is false
for all $\xa(\tau)$. Therefore, we have (\ref{equ:27}).

Following (\ref{equ:27}),

\begin{figure*}[t]
\normalsize \setcounter{equation}{28}
\begin{align}
 & \mathds{E}[\xB(b)]  \leq \left\lceil \frac{4 L H^2 a_{\max}^2 \ln t_2(b)}{
\Delta^2_{\min} } \right\rceil  + \sum\limits_{\tau = 1}^{t_2(b)}
\sum\limits_{s_{h_1} = 1}^{\tau-1} \dots
\sum\limits_{s_{h_{|\xA^*|}} = 1}^{\tau-1}
 \sum\limits_{s_{p_1} = l}^{\tau-1} \dots \sum\limits_{s_{p_{|\xA_{\xa(\beta)}|}} = l}^{\tau-1} \frac{2 H S_{\max}}{\pi_{\min}}
\tau^{-\frac{L \epsilon_{\min}}{28 S^2_{\max} r^{2}_{\max}
 \hat{\pi}^{2}_{\max}}} \label{equ:27}
\end{align}
\setcounter{equation}{31} \hrulefill \vspace*{4pt}
\end{figure*}

\begin{align}
 & \mathds{E}[\xB(b)]  \leq  \frac{4 L H^2 a_{\max}^2 \ln n}{
\Delta^2_{\min} } + 1 \nonumber\\
& +  \frac{H S_{\max}}{\pi_{\min}} \sum\limits_{\tau = 1}^{\infty} 2
\tau^{-\frac{L \epsilon_{\min} - 56 H S^2_{\max} r^{2}_{\max}
\hat{\pi}^{2}_{\max}}{28 S^2_{\max} r^{2}_{\max}
 \hat{\pi}^{2}_{\max}}} \label{equ:lt}\\
& = \frac{4 L H^2 a_{\max}^2 \ln n}{ \Delta^2_{\min} } + 1 + \frac{H
S_{\max}}{\pi_{\min}} \sum\limits_{\tau = 1}^{\infty}
 2\tau^{-2} \label{equ:26}\\
& = \frac{4 L H^2 a_{\max}^2 \ln n}{ \Delta^2_{\min} } + 1 +
\frac{\pi H S_{\max}}{3 \pi_{\min} } \nonumber
\end{align}


(\ref{equ:26}) follows since $L \geq 56(H+1) S^2_{\max} r^{2}_{\max}
\hat{\pi}^{2}_{\max}/\epsilon_{\min}$.

According to (\ref{equ:f1}),
\begin{align}
 & \sum\limits_{\xa: \gamma^{\xa} < \gamma^*} \mathbb{E}[B^\xa(b)]
 = \sum\limits_{i = 1}^{N}
 \mathbb{E}[\xB(b)] \nonumber \\
 & \leq \frac{4 N L H^2 a_{\max}^2 \ln n}{ \Delta^2_{\min} } + N +
\frac{\pi N H S_{\max}}{3 \pi_{\min} } \label{equ:ba}
\end{align}

Note that the total number of plays of arm $\xa$ at the end of block
$b(n)$ is equal to the total number of plays of arm $\xa$ during
SB2s (the regenerative cycles of visiting state $\zeta^{\xa}$) plus
the total number of plays before entering the regenerative cycles
plus one more play resulting from the last play of the block which
is state $\zeta^{\xa}$. So we have
\begin{align}
E[T^{\xa}(n)] \leq \left(\frac{1}{\Pi^{\xa}_{\min}} + M^{\xa}_{\max}
+1\right)E[B^{\xa}(b(n))]. \nonumber
\end{align}

Therefore,
\begin{align}
 & \sum\limits_{\xa: \gamma^{\xa} < \gamma^*} (\gamma^* - \gamma^{\xa}) \mathbb{E}[T^{\xa}(n)] \nonumber \\
 & \leq \Delta_{\max} \sum\limits_{\xa: \gamma^{\xa} < \gamma^*} \left(\frac{1}{\Pi^{\xa}_{\min}} + M^{\xa}_{\max}
+1\right)E[B^{\xa}(b(n))] \\
 & \leq \Delta_{\max} \left(\frac{1}{\Pi_{\min}} + M_{\max} +1
\right) \sum\limits_{\xa: \gamma^{\xa} < \gamma^*} E[B^{\xa}(b(n))] \\
 & \leq Z_1 \ln n + Z_2 \nonumber
\end{align}
where
\begin{align}
& Z_1 = \Delta_{\max} \left(\frac{1}{\Pi_{\min}} + M_{\max} +1
\right) \frac{4 N L H^2 a_{\max}^2}{ \Delta^2_{\min} }, \nonumber\\
& Z_2 =  \Delta_{\max} \left(\frac{1}{\Pi_{\min}} + M_{\max} +1
\right) \left(N + \frac{\pi N H S_{\max}}{3 \pi_{\min} } \right)
\nonumber
\end{align}

\end{IEEEproof}

Now we show our main results on the regret of CLRMR policy as in
Theorem \ref{theorem:2}. 

\begin{theorem} \label{theorem:2} When using any constant $L \geq 56(H+1) S^2_{\max} r^{2}_{\max}
\hat{\pi}^{2}_{\max}/\epsilon_{\min}$,
 the regret of CLRMR can be upper bounded uniformly
over time by the following,
\begin{align}
& \mathfrak{R}^{CLRMR}(n) \leq Z_3 \ln n + Z_4 \label{equ:50}
\end{align}
where
\begin{align}
 & Z_3 = Z_1 + Z_5 \frac{4 N L H^2
a_{\max}^2 }{ \Delta^2_{\min}
 } \nonumber\\
 & Z_4 = Z_2 + \gamma^*(\frac{1}{\pi_{\min}} + M_{\max}
+1) + Z_5 (N + \frac{\pi N H S_{\max}}{3 \pi_{\min} }) \nonumber
\end{align}
and
\begin{align}
 Z_5 = \gamma'_{\max} (\frac{1}{\Pi_{\min}}+ M_{\max} +1 -
\frac{1}{\pi_{\max}}) + \gamma^* M^*_{\max} \nonumber
\end{align}
\end{theorem}

\begin{IEEEproof}
Denote the expectations with respect to policy CLRMR given $\zeta$
by $E_\zeta$. Then the regret is bounded as,
\begin{align}
& \mathfrak{R}^{CLRMR}_\zeta(n) = \gamma^* \mathbb{E}_\zeta[T(n)] - \mathbb{E}_\zeta[\sum\limits_{t=1}^{T(n)} \sum\limits_{i \in \xA_{\xa(t)}} a_i(t) r^{i}_{x_i(t)} ] \nonumber \\
& \quad + \gamma^* \mathbb{E}_\zeta[n-T(n)] - \mathbb{E}_\zeta[\sum\limits_{t=T(n)+1}^{n} \sum\limits_{i \in \xA_{\xa(t)}} a_i(t) r^{i}_{x_i(t)}] \nonumber \\
& \leq \left(\gamma^* \mathbb{E}_\zeta[T(n)] - \sum\limits_{\xa}
 \gamma^{\xa} \mathbb{E}_\zeta[T^{\xa}(n)] \right) + \gamma^*
 \mathbb{E}_\zeta[n-T(n)] \nonumber\\
& \quad + \sum\limits_{\xa} \gamma^{\xa}
  \mathbb{E}_\zeta[T^{\xa}(n)] -  \mathbb{E}_\zeta[\sum\limits_{t=1}^{T(n)} \sum\limits_{i \in \xA_{\xa(t)}} a_i(t) r^{i}_{x_i(t)}
  ] \nonumber\\
& \leq Z_1 \ln n + Z_2 + \gamma^*(\frac{1}{\Pi_{\min}} + M_{\max}
+1) \label{equ:102}\\
& \quad + \left( \sum\limits_{\xa} \gamma^{\xa}
  \mathbb{E}_\zeta[T^{\xa}(n)] -  \mathbb{E}_\zeta[\sum\limits_{t=1}^{T(n)} \sum\limits_{i \in \xA_{\xa(t)}} a_i(t) r^{i}_{x_i(t)}
  ] \right). \nonumber\\
\end{align}
where (\ref{equ:102}) follows from Theorem \ref{theorem:1} and
$\mathbb{E}_\zeta[n-T(n)] \leq \frac{1}{\Pi_{\min}} + M_{\max} +1$.

Note that
\begin{align}
  & \sum\limits_{\xa} \gamma^{\xa} \mathbb{E}_\zeta[T^{\xa}(n)] -  \mathbb{E}_\zeta[\sum\limits_{t=1}^{T(n)}
  \sum\limits_{i \in \xA_{\xa(t)}} a_i(t) r^{i}_{x_i(t)}] \nonumber\\
  & \leq \gamma^* \mathbb{E}_\zeta[T^*(n)] + \sum\limits_{\xa: \gamma^{\xa} < \gamma^*}
  \gamma^{\xa} \mathbb{E}_\zeta[T^{\xa}(n)] \nonumber\\
  & \quad - \sum\limits_{i \in \xA_{\xa^*}} \sum\limits_{y \in S^i}
  a_i^* r^i_y \mathbb{E}_\zeta[\sum\limits_j^{B^*(b(n))} \sum\limits_{Y^i_t \in Y^i(j)} \mathds{1}(Y^i_t = y)
  ] \nonumber\\
  & \quad - \sum\limits_{\xa: \gamma^{\xa} < \gamma^*} \sum\limits_{i \in \xA_{\xa}} \sum\limits_{y \in S^i}
  a_i r^i_y \mathbb{E}_\zeta[\sum\limits_j^{B^{\xa}(b(n))} \sum\limits_{Y^i_t \in Y^i_2(j)} \mathds{1}(Y^i_t = y)
  ] \label{equ:32}
\end{align}
where the inequality above comes from counting only in $Y^i_2(j)$
instead of $Y^i(j)$ in (\ref{equ:32}). Then applying Lemma
\ref{lemma:2} to (\ref{equ:32}), we have
\begin{align}
&  \mathbb{E}_\zeta[\sum\limits_j^{B^{\xa}(b(n))} \sum\limits_{Y^i_t
\in Y^i_2(j)} \mathds{1}(Y^i_t = y) ] =
\frac{\pi^i_y}{\pi^i_{\zeta^i}}  \mathbb{E}_\zeta[B^{\xa}(b(n))].
\nonumber
\end{align}
So
\begin{align}
& - \sum\limits_{\xa: \gamma^{\xa} < \gamma^*} \sum\limits_{i \in
\xA_{\xa}} \sum\limits_{y \in S^i}
  a_i r^i_y \mathbb{E}_\zeta[\sum\limits_j^{B^{\xa}(b(n))} \sum\limits_{Y^i_t \in Y^i_2(j)} \mathds{1}(Y^i_t = y)
  ] \nonumber\\
& \leq - \sum\limits_{\xa: \gamma^{\xa} < \gamma^*}
\frac{\gamma^{\xa}}{\pi_{\max}} \mathbb{E}_\zeta[B^{\xa}(b(n))].
\label{equ:33}
\end{align}
Also note that
\begin{align}
  & \sum\limits_{\xa: \gamma^{\xa} < \gamma^*}
    \gamma^{\xa} \mathbb{E}_\zeta[T^{\xa}(n)] \nonumber\\
  & \leq \sum\limits_{\xa: \gamma^{\xa} < \gamma^*}
    \gamma^{\xa} (\frac{1}{\pi^{\xa}_{\min}} + M^{\xa}_{\max}
    +1) \mathbb{E}_\zeta [B^{\xa}(b(n))] \label{equ:34}
\end{align}

Inserting (\ref{equ:33}) and (\ref{equ:34}) into (\ref{equ:32}), we
get
\begin{align}
  & \sum\limits_{\xa} \gamma^{\xa} \mathbb{E}_\zeta[T^{\xa}(n)] -  \mathbb{E}_\zeta[\sum\limits_{t=1}^{T(n)}
  \sum\limits_{i \in \xA_{\xa(t)}} a_i(t) r^{i}_{x_i(t)}] \nonumber\\
  & \leq \gamma^* \mathbb{E}_\zeta[T^*(n)] \nonumber\\
  & \quad + \sum\limits_{\xa: \gamma^{\xa} < \gamma^*}
    \gamma^{\xa} (\frac{1}{\Pi^{\xa}_{\min}}+ M^{\xa}_{\max} +1 - \frac{1}{\pi_{\max}}) \mathbb{E}_\zeta [B^{\xa}(b(n))] \nonumber\\
  & \quad - \sum\limits_{i \in \xA_{\xa^*}} \sum\limits_{y \in S^i}
  a_i^* r^i_y \mathbb{E}_\zeta[\sum\limits_j^{B^*(b(n))} \sum\limits_{Y^i_t \in Y^i(j)} \mathds{1}(Y^i_t = y)
  ] \nonumber\\
  & = Q^*(n) \nonumber\\
  & \quad + \sum\limits_{\xa: \gamma^{\xa} < \gamma^*}
    \gamma^{\xa} (\frac{1}{\Pi^{\xa}_{\min}}+ M^{\xa}_{\max} +1 - \frac{1}{\pi_{\max}}) \mathbb{E}_\zeta [B^{\xa}(b(n))], \nonumber
\end{align}
where
\begin{align}
& Q^*(n) = \gamma^* \mathbb{E}_\zeta[T^*(n)] \nonumber\\
 & \quad - \sum\limits_{i \in
\xA_{\xa^*}} \sum\limits_{y \in S^i}
  a_i^* r^i_y \mathbb{E}_\zeta[\sum\limits_j^{B^*(b(n))} \sum\limits_{Y^i_t \in Y^i(j)} \mathds{1}(Y^i_t = y)
  ] \nonumber
\end{align}

We now consider the upper bound for $Q^*(n)$. We note that the total
number of time slots for playing all suboptimal arms is at most
logarithmic, so the number of time slots in which the optimal arm is
not played is at most logarithmic. We could then combine the
successive blocks in which the best arm is played, and denote by
$\bar{Y}^*(j)$ the $j$-th combined block. Denote $\bar{b}^*$ as the
total number of combined blocks up to block $b$. Each combined block
$\bar{Y}^*$ starts after dis-continuity in playing the optimal arm,
so $\bar{b}^*(n)$ is less than or equal to total number of completed
blocks in which the best arm is not played up to time $n$. Thus,
\begin{equation}
\mathbb{E}_\zeta[\bar{b}^*(n)] \leq \sum_{\xa: \gamma^{\xa} <
\gamma^*} \mathbb{E}_\zeta[B^{\xa}(b(n))]. \label{equ:35}
\end{equation}

Each combined block $\bar{Y}^*$ consists of two sub-blocks:
$\bar{Y}^*_1$ which contains the state vectors for the optimal arm
visited from beginning of $\bar{Y}^*$ (empty if the first state is
$\zeta^*$) to the state right before hitting $\zeta^*$ and sub-block
$\bar{Y}^*_2$ which contains the rest of $\bar{Y}^*$ (a random
number of regenerative cycles). Denote the length of $\bar{Y}^*_1$
by $|\bar{Y}^*_1|$ and the length of $\bar{Y}^*_2$ by
$|\bar{Y}^*_2|$. We denote $\bar{Y}^i_2(j)$ by the states for Markov
chain $i$ for all $i \in \xA_{\xa^*}$ in $\bar{Y}^*_2$.

Therefore we get the upper bound for $Q^*(n)$ as
\begin{align}
& Q^*(n) = \gamma^* \mathbb{E}_\zeta[T^*(n)] \nonumber\\
 & \quad - \sum\limits_{i \in
\xA_{\xa^*}} \sum\limits_{y \in S^i}
  a_i^* r^i_y \mathbb{E}_\zeta[\sum\limits_j^{B^*(b(n))} \sum\limits_{Y^i_t \in Y^i(j)} \mathds{1}(Y^i_t = y)
  ] \label{equ:37} \\
& \leq \sum\limits_{i \in \xA_{\xa^*}} \sum\limits_{y \in S^i} a_i^*
r^i_y \pi^i_y \mathbb{E}_\zeta[\sum\limits_{j = 1}^{\bar{b}^*(n)
}|\bar{Y}^*_2| ]
\label{equ:38}\\
& \quad  - \sum\limits_{i \in \xA_{\xa^*}} \sum\limits_{y \in S^i}
a_i^* r^i_y \mathbb{E}_\zeta[\sum\limits_{j = 1}^{\bar{b}^*(n) }
\sum\limits_{Y^i_t \in \bar{Y}^i_2(j)} \mathds{1}(Y^i_t = y) ]
\label{equ:39}\\
 & \quad + \sum\limits_{i \in \xA_{\xa^*}} \sum\limits_{y \in S^i} \gamma^* \mathbb{E}_\zeta[\sum\limits_{j = 1}^{\bar{b}^*(n)
}|\bar{Y}^*_1| ] \label{equ:40} \\
& \leq \gamma^* M^*_{\max} \sum\limits_{\xa: \gamma^{\xa} <
\gamma^*} \mathbb{E}_\zeta [B^{\xa}(b(n))] \label{equ:41}
\end{align}
where the inequality in (\ref{equ:38}) comes from counting only the
rewards obtained in sub-block $\bar{Y}^i_2$ in (\ref{equ:37}). Also,
note that based on Lemma \ref{lemma:2}, (\ref{equ:38}) equals
(\ref{equ:39}), and therefore we have the inequality (\ref{equ:41}).

Hence, $\forall \zeta$,
\begin{align}
& \mathfrak{R}^{CLRMR}_\zeta(n) \leq Z_1 \ln n + Z_2 +
\gamma^*(\frac{1}{\pi_{\min}} + M_{\max}
+1) \label{equ:31}\\
& \quad + \sum\limits_{\xa: \gamma^{\xa} < \gamma^*}
    \gamma^{\xa} (M^{\xa}_{\max} +1) \mathbb{E}_\zeta
    [B^{\xa}(b(n))] \nonumber\\
&  \quad + \gamma^* M^*_{\max} \sum\limits_{\xa: \gamma^{\xa} <
\gamma^*} \mathbb{E}_\zeta [B^{\xa}(b(n))]  \nonumber\\
& \leq Z_1 \ln n + Z_2 + \gamma^*(\frac{1}{\pi_{\min}} + M_{\max}
+1)  \nonumber\\
& + (\gamma'_{\max} (\frac{1}{\Pi_{\min}}+ M_{\max} +1 -
\frac{1}{\pi_{\max}}) + \gamma^* M^*_{\max})
\mathbb{E}_\zeta [B^{\xa}(b(n))] \nonumber\\
& \leq Z_3 \ln n + Z_4, \label{equ:50}
\end{align}
where (\ref{equ:50}) follows from Theorem \ref{theorem:1} and
(\ref{equ:ba}), and
\begin{align}
 & Z_3 = Z_1 + Z_5 \frac{4 N L H^2
a_{\max}^2 }{ \Delta^2_{\min}
 } \nonumber\\
 & Z_4 = Z_2 + \gamma^*(\frac{1}{\Pi_{\min}} + M_{\max}
+1) + Z_5 (N + \frac{\pi N H S_{\max}}{3 \pi_{\min} }). \nonumber
\end{align}
$Z_5$ is defined as
\begin{align}
 Z_5 = \gamma'_{\max} (\frac{1}{\Pi_{\min}}+ M_{\max} +1 -
\frac{1}{\pi_{\max}}) + \gamma^* M^*_{\max}. \nonumber
\end{align}
\end{IEEEproof}

Theorem \ref{theorem:2} shows when we use a constant $L \geq 56(H+1)
S^2_{\max} r^{2}_{\max} \hat{\pi}^{2}_{\max}/\epsilon_{\min}$, the
regret of Algorithm \ref{alg:restless} is upper-bounded uniformly
over time $n$ by a function that grows as $O(N^3 \ln n)$. However,
when $S_{\max}$, $r_{\max}$, $\hat{\pi}_{\max}$ or $\epsilon_{\min}$
(or the bound of them) are unknown, the upper bound of regret can
not be guaranteed to grow logarithmically in $n$.

When no knowledge about the system is available, we extend the CLRMR
policy to achieve a regret bounded uniformly over time $n$ by a
function that grows as $O(N^3 L(n) \ln n)$, using any arbitrarily
slowly diverging non-decreasing sequence $L(n)$ in Algorithm
\ref{alg:restless}. Since $L(n)$ could grow arbitrarily slowly, this
modified version of CLRMR, named CLRMR-LN, could achieve a regret
arbitrarily close to the logarithmic order. We present our analysis
in Theorem \ref{theorem:3}.

\theorem\label{theorem:3} When using any arbitrarily slowly
diverging non-decreasing sequence $L(n)$ (i.e., $L(n)\rightarrow
\infty$ as $n \rightarrow \infty$), and replacing
(\ref{equ:maxrestless}) in Algorithm \ref{alg:restless} accordingly
with
\begin{equation}\label{equ:growing}
    \max\limits_{\xa \in \xF} a_i \left(\bar{z}^i_2 + \sqrt{ \frac{ L(n(t_2)) \ln t_2 }{ m^i_2}}
    \right)
\end{equation}
where $n(t_2)$ is the time when total number of time slots spent in
SB2 is $t_2$, the expected regret under this modified version of
CLRMR, named CLRMR-LN policy, is at most
\begin{align}
& \mathfrak{R}^{CLRMR-LN}(n) \leq Z_6 L(n) \ln n + Z_7
\end{align}
where $Z_6$ and $Z_7$ are constants.

\begin{IEEEproof}

Replacing $c_{t, s}$ with $\sqrt{ \frac{ L(n(t)) \ln t }{ s} }$, and
replacing $L$ with $L(n(t_2(b)))$ or $L(n(\tau))$ accordingly in the
proof of Theorem \ref{theorem:1}, (\ref{equ:f1}) to (\ref{equ:lt})
still stand.

$L(n(\tau))$ is a diverging non-decreasing sequence, so there exists
a constant $\tau'$ such that for all $\tau \geq \tau'$, $L(n(\tau))
\geq 56(H+1) S^2_{\max} r^{2}_{\max}
\hat{\pi}^{2}_{\max}/\epsilon^{\min}$, which implies
$\tau^{-\frac{L(n(\tau)) \epsilon^{\min} - 56 H S^2_{\max}
r^{2}_{\max} \hat{\pi}^{2}_{\max}}{28 S^2_{\max} r^{2}_{\max}
 \hat{\pi}^{2}_{\max}}}
\leq \tau^{-2}$.

Thus, we have
\begin{align}
 & \mathds{E}[\xB(b)]  \leq
 \frac{4 L(n(t_2(b))) H^2 a_{\max}^2 \ln n}{ \Delta^2_{\min} } + 1 \\
 & \quad + \frac{H
S_{\max}}{\pi_{\min}} \sum\limits_{\tau = 1}^{\infty}
 2\tau^{-2} + Z_8 \nonumber\\
& \leq \frac{4 L(n) H^2 a_{\max}^2 \ln n}{ \Delta^2_{\min} } + 1 +
\frac{\pi H S_{\max}}{3 \pi_{\min} } + Z_8
\end{align}
where
\begin{equation}
 Z_8 = \frac{H
S_{\max}}{\pi_{\min}}  \sum\limits_{\tau = 1}^{\tau'-1} 2
\tau^{-\frac{L \epsilon^{\min} - 56 H S^2_{\max} r^{2}_{\max}
\hat{\pi}^{2}_{\max}}{28 S^2_{\max} r^{2}_{\max}
 \hat{\pi}^{2}_{\max}}}
\end{equation}

Then we can according have
\begin{align}
 & \sum\limits_{\xa: \gamma^{\xa} < \gamma^*} (\gamma^* - \gamma^{\xa}) \mathbb{E}[T^{\xa}(n)] \nonumber \\
 & \leq Z_9 L(n) \ln n + Z_2 + \Delta_{\max} \left(\frac{1}{\Pi_{\min}} + M_{\max} +1
\right) N Z_8. \nonumber
\end{align}
where
\begin{align}
 Z_9 = \Delta_{\max} \left(\frac{1}{\Pi_{\min}} + M_{\max} +1
\right) \frac{4 N H^2 a_{\max}^2}{ \Delta^2_{\min} }.
\end{align}
So
\begin{align}
& \mathfrak{R}^{CLRMR-LN}(n) \leq Z_6 L(n) \ln n + Z_7,
\end{align}
where
\begin{align}
 & Z_6 =  Z_9 + Z_5 \frac{4 N H^2
a_{\max}^2 }{ \Delta^2_{\min} } \nonumber\\
 & Z_7 = Z_2  + \gamma^*(\frac{1}{\Pi_{\min}} + M_{\max}
   +1) \nonumber\\
 & \quad + \Delta_{\max} \left(\frac{1}{\Pi_{\min}} + M_{\max} +1 \right) N
  Z_7\\
 &  + Z_5(N + \frac{\pi N H S_{\max}}{3 \pi_{\min} } + N Z_7).
\nonumber
\end{align}

\end{IEEEproof}

\section{Applications and Simulation Results} \label{sec:app:simulation}

We now present an evaluation of our policy over stochastic versions
of two combinatorial network optimization problems of practical
interest:  stochastic shortest path (for routing), and stochastic
bipartite matching (for channel allocation).

\subsection{Stochastic Shortest Path} \label{subsection:shortest:path}

In the stochastic shortest path problem, given a graph $G=(V, E)$,
with edge weights $(D_{ij})$ stochastically varying with time as
restless Markov chains with unknown dynamics, we seek to find a path
between a given source $s$ and destination $t$ with minimum expected
delay. We can apply the CLRMR policy to this problem, with some very
minor modifications to the policy and the corresponding regret
definition to be applicable to a minimization problem instead of
maximization.

For the stochastic shortest path problems, each path between $s$ and
$t$ is mapped to an arm. Although the number of paths could grow
exponentially with the number of Markov chains, $|E|$. CLRMR
efficiently solves this problem with polynomial storage $|E|$ and
regret scaling as $O(|E|^3 \log n)$.

\begin{figure}[ht]
\centering
   \includegraphics[width=0.28\textwidth] {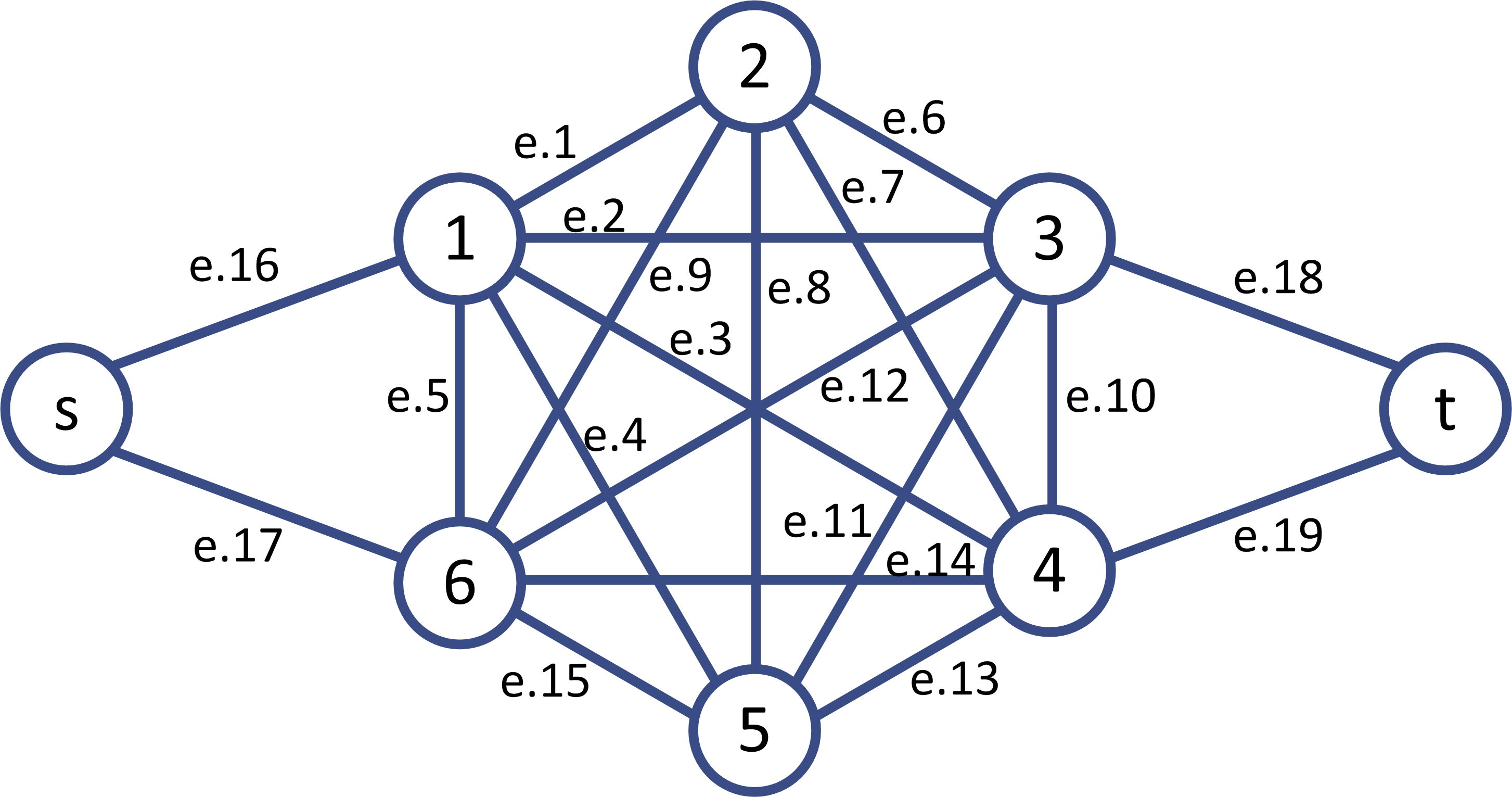}
\vspace{-.2cm}
 \caption{A graph with $19$ links and $260$ acyclic paths between $s$ and $t$ for stochastic
shortest path routing.} \label{fig:shortest:path}
\end{figure}

\begin{figure}[t]
    \centering
        \includegraphics[width=0.34\textwidth]{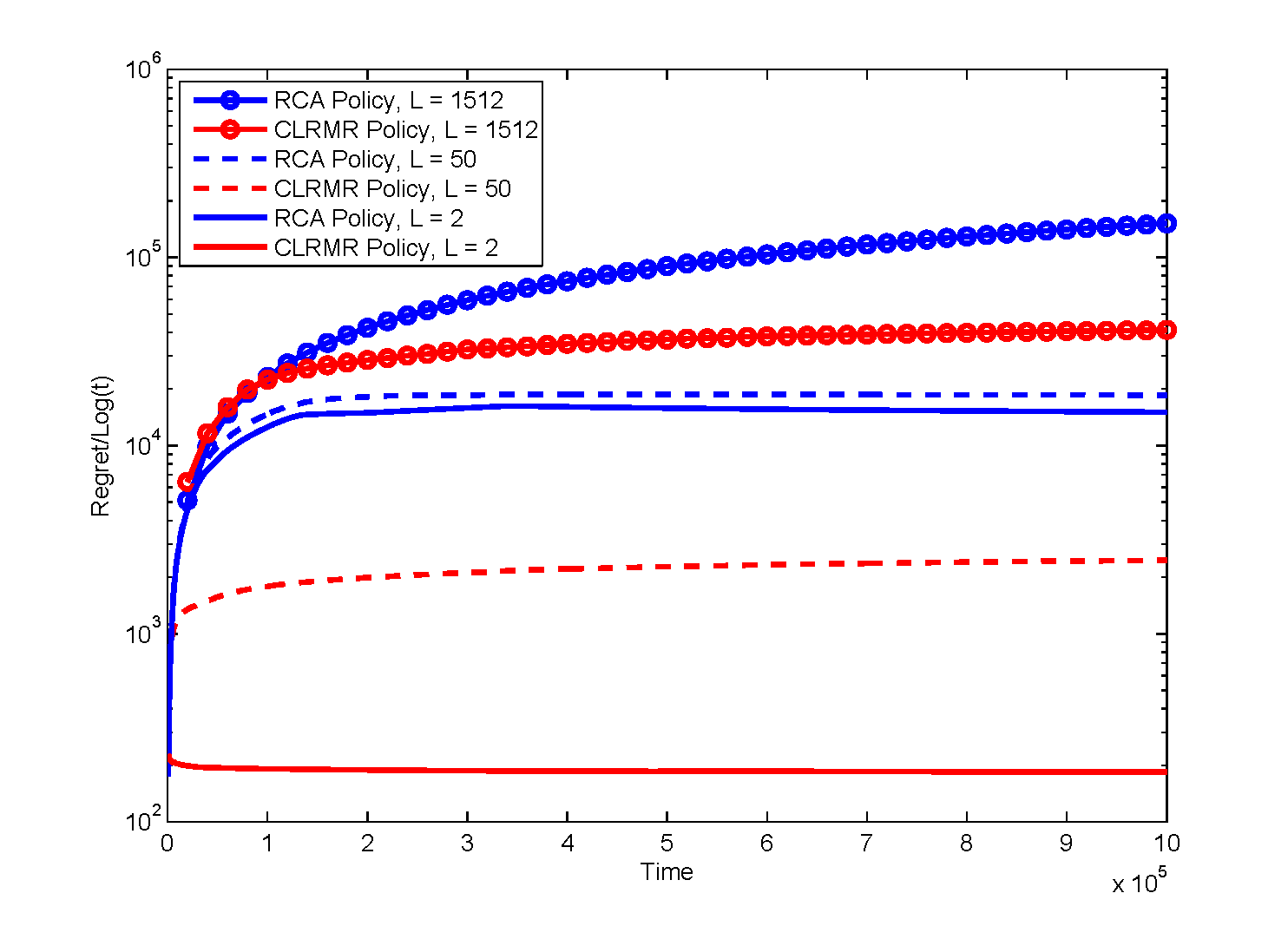}
\vspace{-.4cm}
    \caption{Comparison of normalized regret $\frac{\mathfrak{R}(n)}{\ln n}$ vs. $n$ time slots for the stochastic shortest path problem.}
    \label{fig:path}
\end{figure}

We show the numerical simulation results for the graph in Figure
\ref{fig:shortest:path}. We assume each link has two states with the
delay 0.1 on good links, and 1 on bad links. Table \ref{table:s1b}
summarizes the transition probabilities on each link.

\begin{table}[h]
\begin{center}
    \begin{tabular}{ l|c||l|c||l|c}
    \hline
    Link  & $p_{01}$, $p_{10}$ & Link &   $p_{01}$, $p_{10}$  & Link & $p_{01}$, $p_{10}$ \\ \hline
    e.1 & 0.2, 0.8& e.8 & 0.3, 0.8 & e.15 & 0.1, 0.8  \\ \hline
    e.2 & 0.3, 0.9& e.9 & 0.1, 0.9 & e.16 & 0.8, 0.1 \\ \hline
    e.3 & 0.2, 0.7& e.10 & 0.9, 0.1& e.17 & 0.2, 0.7  \\ \hline
    e.4 & 0.7, 0.1& e.11 & 0.3, 0.8& e.18 & 0.9, 0.1  \\ \hline
    e.5 & 0.3, 0.9& e.12 & 0.2, 0.7& e.19 & 0.3, 0.8 \\ \hline
    e.6 & 0.2, 0.7& e.13 & 0.8, 0.1&  &  \\ \hline
    e.7 & 0.2, 0.8& e.14 & 0.4, 0.8&  &  \\ \hline
    \end{tabular}
\vspace{0.1in} \caption{Transition probabilities}\label{table:s1b}
\end{center}
\end{table}
\vspace{-.4cm}

 Figure \ref{fig:path} shows the simulation results.
We see that our proposed CLRMR performs better than RCA, the
algorithm presented in~\cite{Tekin:restless:infocom} for all $L$
values considered. If we let $L = 1512$ in this problem, we have
that $L \geq 56(H+1) S^2_{\max} r^{2}_{\max}
\hat{\pi}^{2}_{\max}/\epsilon_{\min}$. For lower values of $L$ it is
not guaranteed by the analysis that the algorithms should yield
logarithmic regret. However, numerically, we find that both policies
seem to achieve logarithmic regret, and yield much better regret
performance, even for much smaller $L$ values. It is unclear whether
this can be proved rigorously or whether it is due low probability
events not captured in the simulations.

\subsection{Stochastic Bipartite Matching for Channel Allocation}

As a second application, we consider an application in a cognitive
radio networks where $M$ secondary users interfering with each other
need to be allocated to $Q$ non-conflicting orthogonal channels. We
assume that, due to geographic dispersion, each user may see
different primary user occupancy behavior in each channel. The
availability of spectrum opportunities on each user-channel
combination (i,j) over a decision period is modeled as a restless
two-state Markov chain. It is easy to show that applying CLRMR to
this problem yields storage linear in $MQ$, and a regret bound that
scales as $O(\min\{M, Q\}^2 M Q\log n)$, following Theorem
\ref{theorem:2}.

We show simulation results comparing CLRMR again with RCA for a
system consisting of 9 orthogonal channels, and 5 secondary users.
The transition probability matrix used for these scenarios is
presented in table \ref{table:2b}.
\begin{table}[h]
\begin{center}
    \begin{tabular}{ |@{\hspace{0.6mm}}c@{\hspace{0.6mm}}|@{\hspace{0.6mm}}c@{\hspace{0.6mm}}|@{\hspace{0.6mm}}c@{\hspace{0.6mm}}|@{\hspace{0.6mm}}c@{\hspace{0.6mm}}|@{\hspace{0.6mm}}c@{\hspace{0.6mm}}|@{\hspace{0.6mm}}c@{\hspace{0.6mm}}|@{\hspace{0.6mm}}c@{\hspace{0.6mm}}|@{\hspace{0.6mm}}c@{\hspace{0.6mm}}|@{\hspace{0.6mm}}c@{\hspace{0.6mm}}|@{\hspace{0.6mm}}c@{\hspace{0.5mm}}| }
    \hline
    &ch.1& ch.2& ch.3 & ch.4& ch.5 & ch.6& ch.7 & ch.8 & ch.9 \\ \hline
    u.1&0.5,0.6& 0.2,0.7& 0.2,0.9 & 0.8,0.1& 0.2,0.7 & 0.3,0.7& 0.2,0.9 & 0.2,0.7 & 0.1,0.9 \\ \hline
    u.2&0.3,0.8& 0.1,0.9& 0.2,0.8 & 0.3,0.7& 0.3,0.6 & 0.2,0.8& 0.4,0.7 & 0.2,0.8 & 0.9,0.2 \\ \hline
    u.3&0.8,0.1& 0.2,0.7& 0.3,0.7 & 0.2,0.8& 0.5,0.6 & 0.2,0.7& 0.2,0.7 & 0.2,0.8 & 0.1,0.9 \\ \hline
    u.4&0.3,0.9& 0.2,0.8& 0.2,0.9 & 0.4,0.6& 0.9,0.2 & 0.2,0.9& 0.2,0.9 & 0.2,0.9 & 0.2,0.9 \\ \hline
    u.5&0.5,0.6& 0.2,0.7& 0.3,0.9 & 0.2,0.7& 0.5,0.5 & 0.2,0.7& 0.8,0.1 & 0.3,0.9 & 0.3,0.9 \\ \hline
    \end{tabular}
\vspace{0.1in} \caption{Transition probabilities $p_{01}$, $p_{10}$
for each user-channel pair} \label{table:2b}
\end{center}
\end{table}

\begin{figure}[t]
    \centering
        \includegraphics[width=0.34\textwidth]{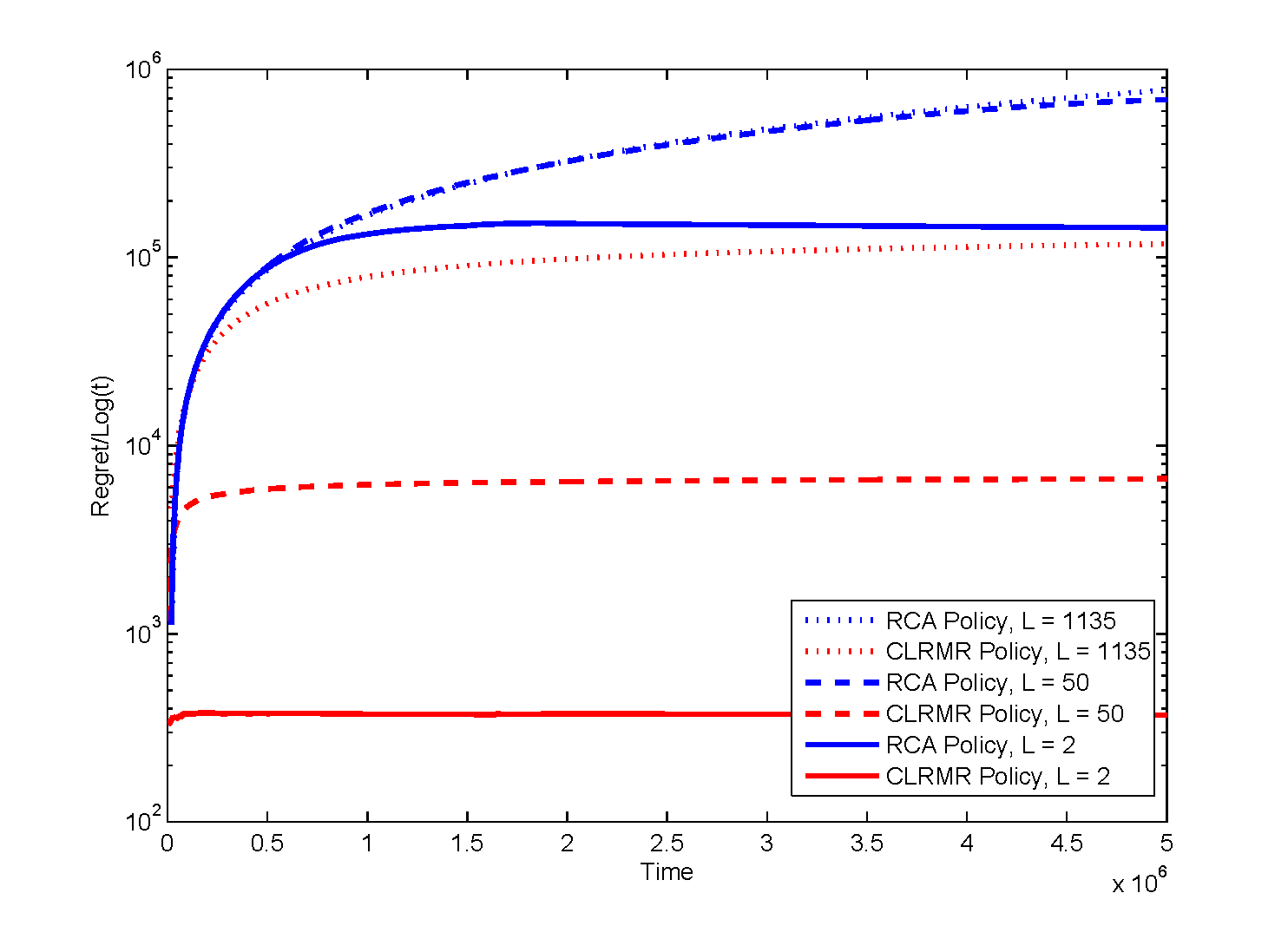}
\vspace{-.4cm}
    \caption{Comparison of normalized regret $\frac{\mathfrak{R}(n)}{\ln n}$ vs. $n$ time slots for Stochastic Bipartite Matching / Channel Allocation Problem.}
    \label{fig:matching}
\end{figure}

\vspace{-.4cm}

The simulation results are shown in Figure \ref{fig:matching}. As in
the stochastic shortest path problem, we find that CLRMR
consistently outperforms RCA, for all values of $L$. Here $L = 1135$
corresponds to ensuring that $L \geq 56(H+1) S^2_{\max} r^{2}_{\max}
\hat{\pi}^{2}_{\max}/\epsilon_{\min}$, which is when the logarithmic
regret is guaranteed in theory. However, again, we see that the
performance seems to improve in practice with smaller $L$ values,
even if it is not be theoretically guaranteed.

\section{Conclusion}\label{sec:conclusion}

We have presented CLRMR, a provably efficient online learning policy
for stochastic combinatorial network optimization with restless
Markovian rewards. This algorithm is widely applicable to many
networking problems of interest, as illustrated by our simulation
based evaluation of the policy over two different problems:
stochastic shortest path and stochastic maximum weight bipartite
matching.

One shortcoming of this work is that our focus has been on designing
and evaluating the policy with respect to the best single-action
policy. However, in general, with restless Markovian rewards, it is
possible to further improve performance by developing an algorithm
that dynamically switches between different actions over time as the
underlying Markov chains evolve. Although this problem is much
harder and remains unsolved except in a special
case~\cite{Dai:icassp}, we hope to investigate it further in our
future work.

\end{document}